\documentclass[runningheads]{llncs}
\usepackage[title]{appendix}
\usepackage{import}
\usepackage{setspace}
\usepackage{graphicx}
\usepackage{amsmath}
\usepackage{algorithm}
\usepackage{algorithmicx}
\usepackage{algpseudocode}
\usepackage[noend]{algcompatible}
\usepackage{longtable}
\usepackage{subfigure} 
\usepackage{wrapfig}
\usepackage{multirow}
\usepackage{color}
\usepackage{bbding}
\usepackage{changes}
\definechangesauthor[name={Per cusse}, color=orange]{per}

\begin{document}
	\title{MUC-driven Feature Importance Measurement and Adversarial Analysis for Random Forest}
	
	\author{Shucen Ma\inst{1} \and Jianqi Shi\inst{1} \and Yanhong Huang\inst{1} \and Shengchao Qin\inst{2}\and Zh\'{e} H\'{o}u \inst{3}}
	\authorrunning{F. Author et al.}
	%
	\institute{National Trusted Embedded Software Engineering Technology Research Center, East China Normal University \and College of Computer Science \& Software Engineering, Shenzhen Univeristy \and School of Information and Communication Technology, Griffith University}
	\maketitle              
	\begin{abstract}
		
		The broad adoption of Machine Learning (ML) in security-critical fields demands the explainability of the approach. However, the research on understanding ML models, such as Random Forest (RF), is still in its infant stage. In this work, we leverage formal methods and logical reasoning to develop a novel model-specific method for explaining the prediction of RF. Our approach is centered around Minimal Unsatisfiable Cores (MUC) and provides a comprehensive solution for feature importance, covering local and global aspects, and adversarial sample analysis. Experimental results on several datasets illustrate the high quality of our feature importance measurement. We also demonstrate that our adversarial analysis outperforms the state-of-the-art method. Moreover, our method can produce a user-centered report, which helps provide recommendations in real-life applications.
		
		
	\end{abstract}
	
	\section{Introduction}
	
	Machine Learning (ML) is ubiquitous nowadays, it has been widely used to perform security and safety-sensitive tasks, such as self-driving \cite{tian2018deeptest}, health care \cite{chen2017disease}, and smart government \cite{alexopoulos2019machine}. However, ML models are widely used as black boxes. That is, it is hard to understand the internal working of the models. The mysterious behavior of ML models is a barrier to the adoption of ML in some applications \cite{molnar2020interpretable}. Consequently, the demand for explainable ML increases. For example, the doctors need to understand how the ML model works before they can comfortably use it in practice. When a patient is diagnosed with an illness, an explainable method can look at important features in the model (used for the positive diagnosis) and analyse whether/how any modification of certain values may lead to a negative diagnosis. Such analysis will likely provide valuable insights in relation to treatment recommendations. 
	
	The research problem we are interested in is explainable AI (XAI), a promising topic in recent years \cite{arrieta2020explainable,gunning2019darpa}. A common way of interpreting ML models is by discovering effective features used in the prediction, namely feature-relevant interpretation. It covers two aspects, including \emph{local explanation}, which focuses on a small region around a sample \cite{Sliwinski_Strobel_Zick_2019,ribeiro2016should}, and \emph{global explanation}, which aims at making the inferential process of a model transparent \cite{vstrumbelj2010explanation,henelius2014peek}. 
	Feature-relevant methods in the literature often just focus on one aspect. They are not comprehensive. Further, most existing feature-relevant tools, like anchor \cite{ribeiro2018anchors} and LIME \cite{ribeiro2016should}, find approximations of the model and use statistical methods to analyse how the input affects the output. They still treat the ML model as a black box, and do not really understand the logic inside the model.  
	
	\emph{Adversarial analysis} is another way to provide insight into the model's behaviour --- it intends to study how a sample can be modified into a different class~\cite{molnar2020interpretable}. The diagnosis and treatment example given above is a suitable application. We are particularly interested in adversarial sample generation, which we use to analyse the reliability of ML models and, in case the predicted class is undesirable, provide recommendations for improving an individual's chance of obtaining desirable results in the future.  
	
	Ensemble trees, such as Random Forest (RF) and boosting, are powerful ML methods, especially for structured data such as spreadsheets and large databases \cite{yu2011unsupervised}. However, explanation methods for them are underdeveloped. Moreover, the non-continuous structure of trees limits the way adversarial analysis can be applied because such structures usually do not have gradients, which are often used to generate adversarial samples~\cite{goodfellow2014explaining,zhang2019generating,papernot2016limitations,moosavi2016deepfool}.

	The semantics of decision trees, especially their variant in the form of binary decision diagrams, are well-understood in the logic and formal methods community. Hence, tree-based ML models are a suitable target for analysis using formal methods, which already play a vital role in the verification of ML models \cite{ehlers2017formal,nie2020varf,chen2019robustness,bride2021silas,zhang2021extracting}. We believe that the application of formal methods in ML may lead to more interesting results and that explanations derived by logical reasoning may well complement the existing statistical interpretation methods. This led to our current work that uses formal methods as the centerpiece to enable various XAI features for ensemble trees. Our main contributions are as follows:
	\vspace{-0.1em}
	\begin{itemize}
		\item We use predicate logic to encode the decision process of ensemble trees into logical formulae. We specifically leverage the Minimal Unsatisfiable Cores (MUC) produced by an SMT solver to compute important features responsible for the prediction of individual instances as local explanations. 
		\item We extend Shapley values
		with the logical information uncovered by MUC, leading to a method for analysing the contributions of features when the model predicts a certain class. 
		\item We improve an existing adversarial attacking algorithm using MUC. Within a similar time, our proposed algorithm generates higher quality adversarial samples in terms of the distance to the original sample.
		\item Our adversarial analysis also allows us to generate a user-centered report that suggests how to improve an individual's chance of being classified as the desired class. Notably, our user-centered report is easy to implement.
		\item We have conducted several experiments and a case study to demonstrate the above points.
	\end{itemize}
	
\begin{figure}[t!]
		\centering
		\includegraphics[width=10cm]{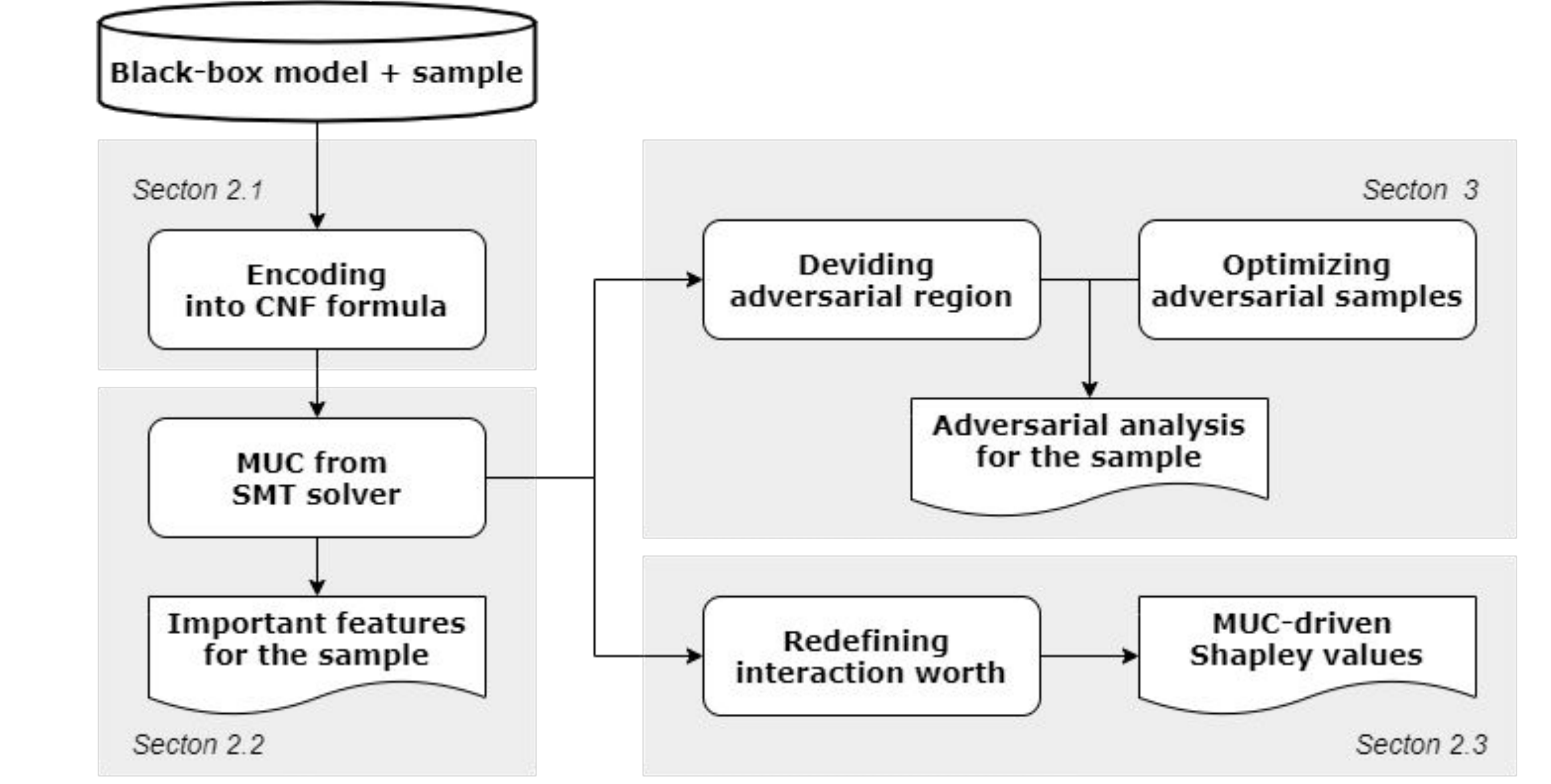}
		\caption{An overview of the proposed approach.}
		\label{fig:flowchart}
	\end{figure}

	In this paper, we focus on Random Forest classifier and binary datasets emphatically. The general workflow of our approach is presented in Figure \ref{fig:flowchart}. The approach consists of four parts: encoding Random Forest into logical formulae (section 2.1), extracting important features in a local analysis, i.e., for specific samples (section 2.2), MUC-driven Shapley values for global feature importance, i.e., for the RF model in general (section 2.3) and adversarial analysis (section 3). Section 4 gives experimental results, followed by related work and conclusion.
	
	\section{MUC-based Feature Importance Analysis}
	
	This section describes our new technique for analyzing feature importance. We have devised a unified MUC-based framework for both local (sample) analysis and global (model) analysis.  
	
	\subsection{Encoding of Random Forest}
	
	For logical analysis, we encode the tree-ensembled Random Forest (RF) into Boolean formulae. We start by encoding one single path in a tree. A decision tree maps the input $\boldsymbol{x}\in{R}^{d}$ to an output, where $\boldsymbol{x} = <x_1,\cdots, x_d>$. Further, a path includes one leaf node \(l\) and several decision nodes \(n \in N_l\), where \(N_l\) is a set including the root node \(n_0\) and the internal nodes between \(n_0\) and \(l\). Each decision node works by comparing the value \(x_i\) of the $i$th feature with the threshold \(\eta_{p_n}\) of this node. The path can be defined as follows formally:
	
	\begin{equation}
		\pi\left( l\right) ::= \underset {n \in N_l}{\bigwedge} {\left(
			\begin{aligned}
				L_{p_n}=n \rightarrow x_i \leq \eta_{p_n} \\
				R_{p_n}=n \rightarrow x_i > \eta_{p_n}
			\end{aligned}
			\right)} \wedge \left(w = v_l \right)   
	\end{equation}
	
	If the node \(n\) is the left child \(L_{p_n}\) of its parent node $p_n$, the parent node holds the condition that \(x_i \leq \eta_{p_n}\). If it is the right child  \(R_{p_n}\), the parent node holds \(x_i > \eta_{p_n}\). And the valuation variable  \(w\) constrains the leaf value \(v_l\) \cite{chen2019robustness}. The path formula \(\pi\left( l\right)\) represents a path from the root to one of the leaves in the tree and its computed leaf value. Then we can encode a tree $t$ as below:
	\vspace{-0.5em}
	\begin{equation}
		\Pi \left(t \right) ::= \underset {l \in L}{\bigvee} \pi \left( l \right)
	\end{equation}
	\vspace{-1em}
	
	\noindent where \(\Pi \left(t \right)\) is the disjunction of path formulae. Further, the encoding of Random Forest contains the conjunction of trees making up the forest and the forest's output. It is usually assumed that RF is a set of $k$ trees  \( \left<t_{1},...,t_{k} \right>\). Given an input $\boldsymbol{x}\in{R}^{d}$, each tree outputs a set of possibilities for each predicted class. Then the forest outputs the class with the highest mean possibility across all trees. Let \(t_{j}\left( \boldsymbol{x} \right)\) be the output of tree \(t_{j}\) and \(t_{j}^{i}\left( \boldsymbol{x} \right)\) be the \(i\)th possibility. The output predicted class of Random Forest can be defined as: 
	
	\begin{equation}
		C\left( \boldsymbol{x}\right) = \frac{1}{k} \underset {i}{\arg\max}  \sum_{j=1}^{k} t_{j}^{i}\left( \boldsymbol{x} \right)    
	\end{equation}
	
	Finally, Random Forest can be encoded as follows:
	
	\begin{equation}
		R\left(\boldsymbol{x} \right) ::= \overset {k}{\underset {j=1}{\bigwedge}} \Pi \left(t_j \right)\wedge \left( output=\frac{1}{k} \underset {i}{\arg\max} \sum_{j=1}^{k}  t_{j}^{i}\left( \boldsymbol{x} \right) \right) 
	\end{equation}
	
	\subsection{Extracting Important Features for Local Analysis}
	
	Nie et al. \cite{nie2020varf} use formal methods to verify the robustness of Random Forest. They transform the robustness property into logical formulae in conjunctive normal form (CNF). The formulae are input into an SMT solver that utilizes the minimal unsatisfiable core (MUC) \cite{barrett2018satisfiability} to verify the robustness of a given instance. 
	
	To explore what features matter in prediction for individual samples to provide a local interpretation, we are inspired to transform the decision process into logical formulae, and extract the influential features according to the MUC by the above method. MUC is formally defined in \cite{nie2020varf} as follows:
	\vspace{-0.5em}
	
	\begin{definition}[Minimal Unsatisfiable Cores]
		Let \(F\) be a CNF formula and \(F_C\) be the set of conjuncts in \(F\),
		\(S \subseteq F_C\) is a \(MUC\) of \(F\) iff whenever \(F\) is
		unsatisfiable, \(S\) is unsatisfiable, and there is no
		\(S'' \subset S\) that is also a MUC.
	\end{definition}
	\vspace{-0.5em}
	
	Every time a CNF formula is unsatisfiable, the DPLL engine \cite{marques2009conflict} will backtrack to search MUC, denoting the minimal conjuncts that make the entire formula unsatisfiable. A noteworthy fact is that MUC is not unique. For efficiency, there is no need for all of them, so we just get one. An example of a CNF formula is:
	\vspace{-0.1em}
	\begin{equation}
		\begin{aligned}
			\varphi &=\omega_1\wedge \omega_2\wedge\omega_3\wedge\omega_4\\ 
			&=\left(x>1\right)\wedge \left(y<0\right)\wedge \left(y>x\right)\wedge \left(y>-3\right)
		\end{aligned}
	\end{equation}
	
	The formula \(\varphi\) is unsatisfiable. The first three clauses combined are inconsistent, but any two of them are consistent. So by the definition above, the DPLL engine will produce a \(MUC=\left(\omega_1,\omega_2,\omega_3\right)\). 
	
	We encode the decision process of Random Forest into a CNF formula:
	
	\begin{equation}
		\Phi_0::=R \left(\boldsymbol{x}\right) \wedge \left(\boldsymbol{x}=\boldsymbol{x}^{org}\right)\wedge \left(output == {y^{org}}\right)
	\end{equation}
	
	The decision process functions as: the assignment $\boldsymbol{x}=\boldsymbol{x}^{org}$ is input into the RF model $R\left(\boldsymbol{x}\right)$ defined by formula (4). After calculation, the model outputs its prediction $output$. When $output$ equals (denoted by $==$) the expected label ${y^{org}}$, the decision process is accurate. The satisfiability of $\Phi_0$ represents the accurate decision process. In order to acquire MUC, notably, we reconstruct the decision process formula in a way resembling `double negation is positive' as follows:
	
	\begin{equation}
		\Phi_1::=R \left(\boldsymbol{x}\right) \wedge \left(\boldsymbol{x}=\boldsymbol{x}^{org}\right)\wedge \left(output \neq {y^{org}}\right)
		\label{formula:keyfeature}
	\end{equation}
	
	In detail, the assignment is:
	
	\begin{equation}
		(\boldsymbol{x}=\boldsymbol{x}^{org})::=\overset {d}{\underset {i=1}{\bigwedge}} x_i=x_i^{org} 
	\end{equation}
	
	The output of a correct prediction should be ${y^{org}}$. The inequality we set in \(\Phi_1\) yields unsatisfiability. Thus, the unsatisfiability of \(\Phi_1\) represents the accurate decision process, and is equivalent to the satisfiability of \(\Phi_0\). Moreover, as the model $R(\boldsymbol{x})$ is certain, it is the assignment of $\boldsymbol{x}$ that results in this contradiction in $\Phi_1$. Thereby, investigating the reason in assignment for the unsatisfiability of \(\Phi_1\) amounts to investigating why \(\Phi_0\) is satisfiable. As a first step, we set the engine to only backtrack to the conjuncts in the assignment. The returned MUC will tell us which features, related to the feature values in the assignment, influence the prediction most. Such straightforward transformation avoids the discretization of continuous features and statistical methods used when drawing features. As a result, our proposed explanation based on logical reasoning ensures precision and integrity.

	\subsection{MUC-driven Shapley values for Global Analysis}
	
	In addition to finding what features matter in individual prediction, we are also interested in the important features of the model for further valuable insight.
	
	Feature importance refers to techniques that assign scores to input features based on how useful they are to the model's prediction globally. Shapley values \cite{shapley1953value} are a popular method from coalitional game theory and are usually used to evaluate feature contributions. Assuming that each feature is a `player' and the prediction is the `payout', Shapley values distribute payout to players according to their contributions to the total payout. The Shapley value is defined as follows:
	
	\begin{equation}
		\phi_i = \sum_{S \subseteq  F \backslash \left\{f_i\right\}} \frac{|S|!\left(|F|-|S|-1\right)!}{|F|!} \left(\Omega(S\cup\left\{f_i\right\})-\Omega\left(S\right)\right)
	\end{equation}
	
	Considering that features often interact in practice, the Shapley value $\phi_i$ of feature $f_i$ is assigned with the weighted average sum of all interactions between $f_i$ and any possible subset $S \subseteq F\ \backslash\ \left\{f_i\right\}$, where $F$ is the set of all features. To be concrete, the interaction is the difference between the \textbf{worth} $\Omega$ of set $S$ with $f_i$ present and the one with $f_i$ withheld, namely $\Omega(S\cup\left\{f_i\right\})-\Omega\left(S\right)$. The classical worth refers to the prediction value $\hat{f_S}\left(S\right)$ output by model $\hat{f_S}$ retrained with subset $S$. In this case, the classical Shapley values analyse feature contributions for all predicted classes of the model output. Our above MUC-based method allows us to define the worth in the following innovative and reasonable form,  enabling the Shapley values to analyse for one specific class globally:
	
	\begin{equation}
		\Omega\left(S\right)=\sum_{j=1}^m {\iota_j},\ \iota_j = 
		\begin{cases}
			1 & S \subseteq MUC^1_{\boldsymbol{x}_j} \wedge y_j = c\\
			-1 & S \subseteq MUC^1_{\boldsymbol{x}_j} \wedge y_j \neq c\\
			0 & S \not\subseteq MUC^1_{\boldsymbol{x}_j}
		\end{cases}
	\end{equation}
	\noindent
	where $c$ is the class of interest, and $MUC^1_{\boldsymbol{x}_j}$ denotes the MUC computed by $\Phi_1\left(\boldsymbol{x}_j\right)$, defined by formula (\ref{formula:keyfeature}). All MUCs are traversed. If $MUC^1_{\boldsymbol{x}_j}$ includes $S$ and $y_j$ equals class $c$, one positive gain is added to the worth, signifying that $S$ has a beneficial influence on this prediction. The other two conditions are similar. Then the worth of set $S$ in our approach depends on the contributions of the MUC, including $S$, to the prediction. In order to further accelerate the calculation, we adopt the sampling method \cite{vstrumbelj2014explaining} to approximate the Shapley values of individual features. Our algorithm is given in Algorithm~\ref{alg:shap}.
	
	\begin{algorithm}[t!]
		\caption{The approximated MUC-driven Shapley values of class $c$}
		\begin{algorithmic}[1]
			\REQUIRE {Feature set ${F}$, all pair of input ({$\boldsymbol {x}_j$},{$y_j$}), class $c$ and all MUCs.}
			\STATE \textbf{function} Calculating Shapley values $\left(F, M\right)$
			\FORALL {$f_i\in {F}$}
			\STATE $\phi_i \leftarrow 0$
			\FOR{ $ \_ = 0, 1, 2, \cdots, M$}
			\STATE $S \leftarrow $ select a random subset from $F \backslash \left\{f_i\right\}$
			\STATE $\phi_i \leftarrow \phi_i +  \displaystyle\frac{|S|!\left(|F|-|S|-1\right)!}{|F|!} \left(\Omega(S\cup\left\{f_i\right\})-\Omega\left(S\right)\right)$
			\ENDFOR
			\STATE add $\phi_i$ to M-Shapley
			\ENDFOR
			\STATE \textbf{return} M-Shapley
			\STATE \textbf{function} Calculating worth $\Omega \left(S, \boldsymbol{x}_j, y_j, MUC, c\right)$
			\STATE $\Omega \leftarrow 0$
			\FORALL {$\boldsymbol{x}_j$}
			\IF {$S \subseteq MUC_{\boldsymbol{x}_j}$}
			\STATE $gain \leftarrow $ if $(y_j = c)$ ? $1$ : $-1 $
			\STATE $\Omega \leftarrow \Omega +$ $gain$
			\ENDIF
			\ENDFOR
			\STATE \textbf{return} $\Omega$
			
		\end{algorithmic}
		\label{alg:shap}
	\end{algorithm}
	
	To be adaptive, the number of iterations $M$ should balance the efficiency, or it can be set to be close to $2^{|F|}$ to decrease the approximation error conditional to permitted time. Our proposed MUC-driven Shapley values, or shortly `M-Shapley',  reserve fairness \cite{molnar2020interpretable} from the interactions in game theory. Moreover, they fully utilize the internal logic of the model that the MUC represents.
	
	\section{MUC-based Adversarial Sample Analysis}

	Local explanations based on statistics and approximation are not scalable enough to solve other relevant issues. For example, as the characteristic of samples, the key features may also expose the defect, where the values of these features can be modified slightly to change its prediction. Theoretically, \emph{adversarial samples} represent these modified samples that make the model output a different prediction. Next, we will show how to utilize our encoded logical solver to generate optimized adversarial samples with the expected predicted class.
	\vspace{-0.8em}
	\paragraph{Dividing adversarial region. }  Suelflow et al. transform the circuit into an SAT instance and use unsatisfiable cores to debug the errors inside~\cite{suelflow2008using}. By iteratively modifying unsatisfiable cores, which define a procedure called `breaking the core', the unsatisfiable SAT instance can turn to be satisfiable, and the circuit recovers. Their findings suggest that, with the guidance of MUC, we may alter the input of the individual sample to change its classification to be expected.
	
	First, we recall the formula \(\left(\ref{formula:keyfeature}\right)\) and use it to get the MUC, from which we obtain what assignment of features results in the original prediction. Then we attempt to break the core. Instead of altering \(\boldsymbol{x}^{org}\) directly, we examine whether there are adversarial samples in the neighborhood. For binary datasets, the expected class of adversarial samples amounts to the class different from the original class. Then formula \(\left(\ref{formula:keyfeature}\right)\) can be transformed into the following form:
	
	\begin{equation}
		\Phi_2::=R \left(\boldsymbol{x}\right)\wedge\sigma\left(\boldsymbol{x},\boldsymbol{x}^{org},\boldsymbol\tau\right) \wedge\left(output \neq {y^{org}}\right) 
		\label{formula:adv}
	\end{equation}
	
	\noindent where
	\begin{equation}
		\sigma\left(\boldsymbol{x},\boldsymbol{x}^{org},\boldsymbol\tau \right)::=\overset {d}{\underset {i=1}{\bigwedge}}
		{ \left| {x_i-x^{org}_i} \right| }\leq \tau_i
	\end{equation}
	
	The search scope denoted by \(\boldsymbol\tau\) is the area around ${\boldsymbol{x}}^{org}$. We want to find adversarial samples inside. The satisfiability of $\Phi_2$ means that there exist adversarial samples around ${\boldsymbol{x}}^{org}$ and when they are input into the RF model $R(\boldsymbol{x})$ for calculation, the model's prediction $output$ does not equal the original label $y^{org}$. Notably, \(\tau_i\) is initialized to zero, and afterwards, it is enlarged iteratively. If feature \(f_i\) is not in the MUC in the current iteration or can not be changed artificially in reality like age, \(\tau_i\) remains unchanged. Otherwise, it increases with a fixed step size. With the guidance of MUC, the search scope expands in higher efficiency. After the first iteration, the search scope may not be large enough to cover adversarial samples. In other words, $\Phi_2$ is still unsatisfiable due to its MUC, which may be different from that last time because the parameters vary and the formula is different, though the form stays the same. Thus we persist in expanding the search scope until no MUCs remain in $\Phi_2$. Correspondingly, $\Phi_2$ is satisfiable, and we find adversarial samples around $\boldsymbol{x}^{org}$. Meanwhile, the search scope touches the field of another class. We name the intersection region `adversarial region'.
	
	\begin{figure}[t!]
		\centering
		\includegraphics[height=2.8cm]{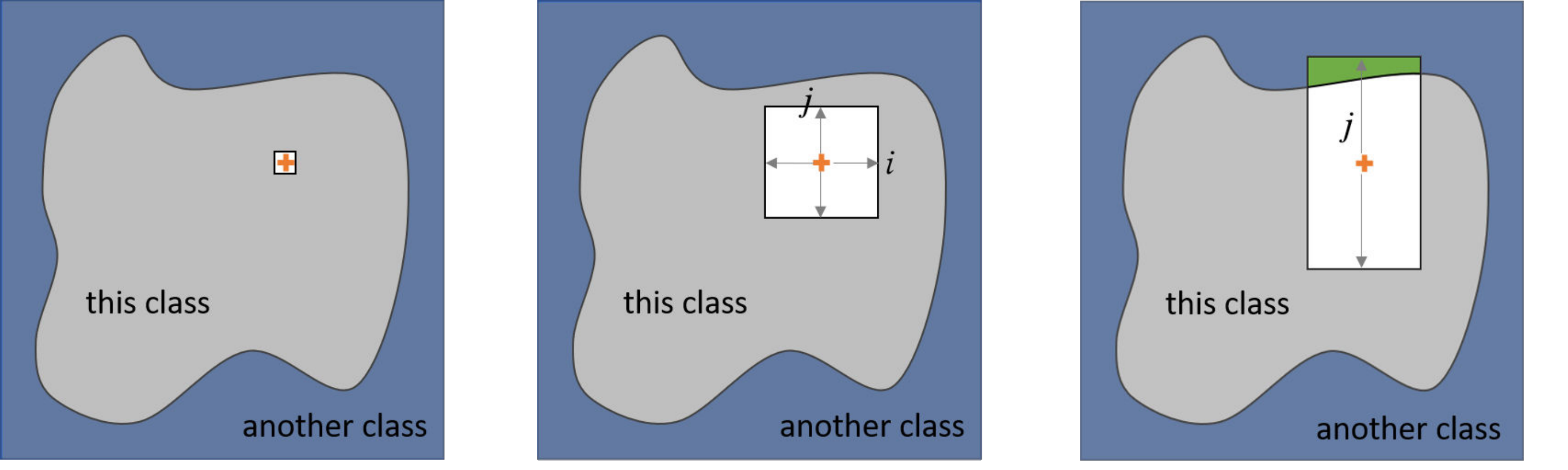}
		\caption{The searching process guided by MUC. (a) In the beginning, the search scope focuses on the initial input. (b) In this iteration, the \(i\)th and \(j\)th feature are in MUC, so the search scope expands on these two components. (c) In this iteration, \(i\)th feature is not in MUC, so the scope on this component stays still. The gray area is the part not yet searched. The white area is the opposite. The green part is the adversarial region.}
	\end{figure}
	\vspace{-0.8em}
	
	\paragraph{Optimizing adversarial samples. }After the above steps, we get the adversarial region full of adversarial samples near the original input. Next, we intend to find the nearest one inside. We propose to solve it through the Zero Order Optimization (Opt-attack) \cite{cheng2018query}. It is used to generate adversarial samples and can be applied to discrete ML models like tree-based models. The objective function is defined below:
	
	\begin{equation}
		g\left(\boldsymbol\theta\right)=\operatorname{argmin}_{\lambda>0} \left( \hat{f} \left(\boldsymbol {x} ^ {org} + \lambda \frac {\boldsymbol\theta}{\left \|\boldsymbol\theta  \right \|}\right) \neq {y^{org}} \right)
	\end{equation}

	\noindent where $\hat{f}$ is the Random Forest function, \(\boldsymbol \theta\) by $\boldsymbol{x}^{adv}-\boldsymbol{x}^{org}$ represents the search direction from some adversarial sample $\boldsymbol{x}^{adv}$ to $\boldsymbol {x}^{org}$ and $\lambda$ represents the distance between. $g(\boldsymbol\theta)$ gets an adversarial sample in the smallest distance along the search direction. Among all the nearest adversarial samples along the respective direction, we want to get the optimal one globally. It is equivalent to the following optimization problem:
	\vspace{-0.2em}
	\begin{equation}
		\underset {\boldsymbol\theta} {\operatorname{min}} \ g\left(\boldsymbol\theta\right)
	\end{equation}
	
	In detail, the Randomized Gradient-Free method in Opt-attack is used to optimize a given search direction. The gradient in each iteration is estimated by:
	\begin{equation}
		\boldsymbol{\hat g} = \frac{g\left( \boldsymbol\theta'\right)- g\left( \boldsymbol\theta \right)}{\beta} \cdot \boldsymbol {u}
	\end{equation}
	
	\noindent where $\boldsymbol\theta' = \boldsymbol\theta + \beta \boldsymbol{u}$  is the perturbed $\boldsymbol\theta$ by a random Gaussian vector $\boldsymbol{u}$ and a smoothing parameter \(\beta > 0\). Then $\boldsymbol \theta$ is updated by $\boldsymbol\theta \leftarrow \boldsymbol{\theta}-\eta\boldsymbol {\hat g}$ with a step size $\eta$ to get optimized. Specifically, the gradient here is just based on the computation of the function values, independent of the non-continuous structure of RF.
	
	\begin{algorithm}[h]
		\caption{Fine-grained and binary search}
		\begin{algorithmic}[1]
			\REQUIRE RF classifier \(\hat{f}\), search direction $\boldsymbol\theta$
			\STATE $\boldsymbol \theta \leftarrow \boldsymbol \theta / \left\| \boldsymbol\theta \right \|$ {\color{blue} \hfill {$\triangleright$ The normalization of $\boldsymbol\theta$} }
			\STATE $v_{out} \leftarrow \left\| \boldsymbol\theta \right \|$, $v_{in} \leftarrow \left\| \boldsymbol\theta \right \|$
			\WHILE{$\hat{f}\left(\boldsymbol{x}_{org}+v_{in}\boldsymbol{\theta}\right) = \hat{f}\left(\boldsymbol{x}_{adv}\right)$}
			\STATE $v_{out} \leftarrow v_{in}$, $v_{in} \leftarrow v_{out} \left(1-\alpha\right)$
			\ENDWHILE
			\WHILE{$v_{out} - v_{in} > \epsilon$}
			\STATE $v_{mid} \leftarrow \left( v_{out}+v_{in}\right) / 2$
			\IF{$\hat{f}\left(\boldsymbol{x}_{org}+v_{mid}\boldsymbol{\theta}\right) = \hat{f}\left(\boldsymbol{x}_{adv}\right)$}
			\STATE $v_{out} \leftarrow v_{mid}$
			\ELSE
			\STATE $v_{in} \leftarrow v_{mid}$
			\ENDIF
			\ENDWHILE
			\State \textbf{return} $v_{out}$  
			{\color{blue} \ \ $\triangleright$ $v_{out} \boldsymbol\theta$ is the final closer adversarial sample in this algorithm}
		\end{algorithmic}
		\label{alg:fine}
		\vspace*{-1mm}
	\end{algorithm}
	
	\begin{algorithm}[ht]
		\caption{Generating the optimized adversarial sample for \(\boldsymbol{x}^{org}\)}
		\begin{algorithmic}[1]
			\REQUIRE RF classifier $\hat f$, CNF of RF \(R(\boldsymbol{x})\), original input \(\boldsymbol{x}^{org}\) and \(y^{org}\),the feature set \(F=\left\{ f_i|0\leq i \leq d\right\}\)
			\STATE \(\boldsymbol\tau \leftarrow \boldsymbol{0}\) 
			\STATE \(\Phi_2\left(x\right) \leftarrow R \left(\boldsymbol{x}\right) \wedge \sigma \left(\boldsymbol {x}, \boldsymbol{x}^{org},\boldsymbol\tau\right) \wedge\left(output \neq {y^{org}}\right)\) 
			\WHILE {$UNSAT=solver\left(\Phi_2\right)$}
			\FORALL{\(f_i\in F\) and $f_i \in$ MUC$^2_{\boldsymbol{x}^{org}}$}
			\STATE \(\tau_i \leftarrow \tau_i + \kappa_i\) 
			\ENDFOR
			\ENDWHILE
			\STATE \textbf {floor} $\leftarrow \boldsymbol {x} ^ {org} - \boldsymbol\tau$, \textbf{ceil} $\leftarrow \boldsymbol {x} ^ {org} + \boldsymbol\tau$
			\STATE Generate random samples in $[$\textbf{floor}, \textbf{ceil}$]$ and gather them in set $X$ 
			\STATE $\lambda_{min} \leftarrow$ a very large number
			\FORALL {$\boldsymbol{x} \in X$ and $\hat{f} \left(\boldsymbol{x}\right) \neq y^{org}$} {\color{blue} \hfill {$\triangleright$ Determining the initial $\theta_0$ among}}
			\STATE $\boldsymbol{\theta} \leftarrow \boldsymbol{x} - \boldsymbol{x}_{org}$ {\color{blue} \hfill { all candidates in adversarial region}}
			\STATE $\lambda = $ Fine-grained and binary search $\left(\hat f,\boldsymbol{\theta}\right)$
			\IF {$\lambda_{min} > \lambda$}
			\STATE $\lambda_{min} \leftarrow \lambda$, $\boldsymbol{\theta}_0 \leftarrow \boldsymbol{\theta}$
			\ENDIF
			\ENDFOR
			\FOR {\(t=0,1,2,\cdots,T\)}\ \  {\color{blue} \hfill{$\triangleright$ Zero order optimization}}
			\STATE\(\boldsymbol\theta_t' = \boldsymbol\theta_t + \beta \boldsymbol{u}_t \cdot \boldsymbol \mu\)
			\STATE Evaluate \(g\left(\boldsymbol\theta_t\right)\) and \(g\left(\boldsymbol\theta_t'\right)\) using fine-grained and binary search
			\STATE \(\boldsymbol{\hat g} \leftarrow \displaystyle\frac{g\left( \boldsymbol\theta_t'\right)- g\left( \boldsymbol\theta_t\right)}{\beta} \cdot \boldsymbol{u}_t\)
			\STATE \(\boldsymbol\theta_{t+1} = \boldsymbol\theta_t -\eta_t \boldsymbol{\hat g}\)
			\ENDFOR
			\State \textbf{return}  \(\boldsymbol{x}^{org}+g\left(\boldsymbol\theta_T\right){\left \|\boldsymbol\theta_T \right \|}\)
		\end{algorithmic}
		\label{alg:wholeprocess}
		\vspace*{-1mm}
	\end{algorithm}
	The initial direction \(\boldsymbol\theta_0\) is required to execute the zero order optimization. To this end, for a given $\boldsymbol{x}^{adv}$ inside the adversarial region, we do a fine-grained search and a binary search to push it to be closer to the boundary along \(\boldsymbol\theta_0\). The corresponding algorithm is presented in Algorithm \ref{alg:fine}, which possesses the distance $v$ between \(\boldsymbol{x}^{adv}\) and \(\boldsymbol{x}^{org}\), $\alpha$ is the increase/decrease ratio, and $\epsilon$ is the stopping tolerance. In the first stage, we search gradually to ensure that the boundary is in $\left[ \boldsymbol{x}_{org}+v_{in}\boldsymbol\theta, \boldsymbol{x}_{org}+v_{out}\boldsymbol\theta \right]$. In the second stage, we conduct a binary search to let the adversarial sample be very close to the boundary.
	
	The whole procedure of generating optimized adversarial samples is summarized in Algorithm \ref{alg:wholeprocess}, where $\kappa_i$ is the step size of enlarging $\tau_i$, and $\boldsymbol{\theta}_T$ is the final optimized search direction. We write $MUC^2_{\boldsymbol{x}^{org}}$ for the MUC computed by $\Phi_2\left(\boldsymbol{x}^{org}\right)$ in each iteration. In particular, except for images, the values of all features may not be in the same order of magnitude. The impact of \(u_t\) on every component of \(\boldsymbol{\theta}_t\) varies too. So we set the vector \(\boldsymbol\mu\) to keep the extent of alteration on every component the same, and the perturbed \(\boldsymbol \theta_t'\) will not go far from \(\boldsymbol\theta_t\) to achieve optimization.
	
	There are several implementation details when applying this algorithm. First, when doing the fine-grained search, the components that never appear in the MUC are not in consideration. The values of them are always the initial. It helps narrow the feature space and increases efficiency. Second, the step size $\kappa_i$ varies according to the order of magnitude of the feature values. And it is best to set the step size as small as possible to ensure that the adversarial region is accurate. 
	
	\section{Experiment and Case Study}
	
	This section is devoted to our experimental results.
	\subsection{Dataset and Setup}
	
	Experiments are carried out with four UCI datasets: credit (binary), breast (binary), MNIST (multiple classes) and heart (binary). We also select two datasets that focus on loans: lending (binary) \cite{ribeiro2018anchors} and bank loan (binary) \cite{personalloanmodel}. The MNIST dataset is specially selected to visualize local explanations and adversarial analysis. Each dataset is cut into two subsets: 80\% for training and 20\% for testing.
	
	We also choose a case study of analysing loan cases. It shows how we provide advice for those who suffer from rejected loan cases. The advice is helpful for them to apply a successful application in the future, where we suggest how to change their submitted information effectively. Furthermore, from the users' perspective, we hope that they do not need to take a huge step out of their comfort zone, which means that the changes had better be as small as possible. Our algorithms can maximize users' advantages on this issue.
	
	We use sklearn to train the Random Forest model based on Python 3.7.x, and use Z3 \cite{de2008z3}  as the underlying SMT solver. Experiments were conducted on a machine with an Intel Core i5-8265U CPU and 16GB RAM.
	
	\subsection{Experimental Results}
	
	\subsubsection{Local feature importance.}
\begin{figure}[t!]
		\centering
		\subfigure[\emph{lending}]{
			\includegraphics[width=3.6cm]{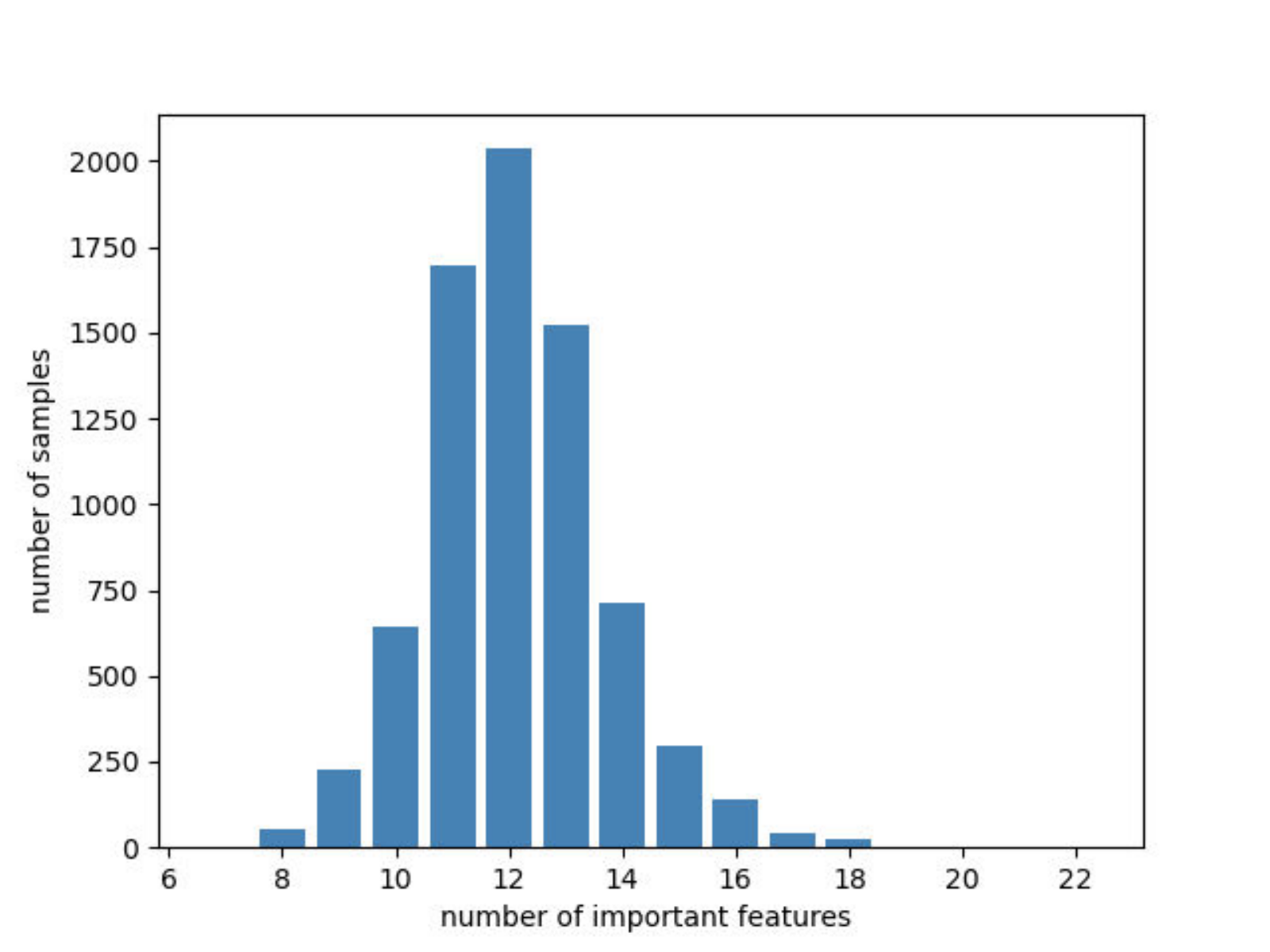}
		}
		\subfigure[\emph{bank loan}]{
			\includegraphics[width=3.6cm]{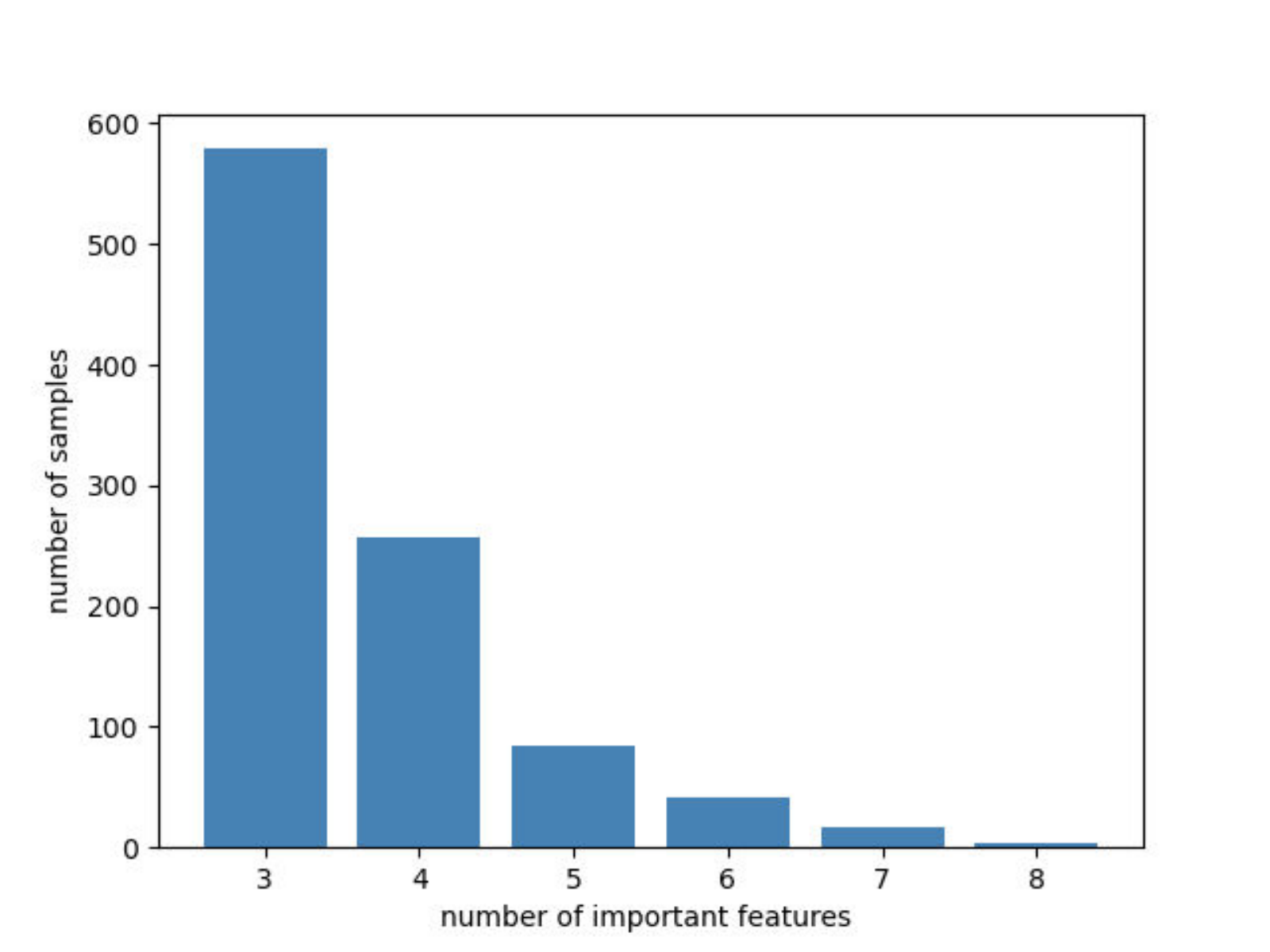}
		}
		\subfigure[\emph{credit}]{
			\includegraphics[width=3.6cm]{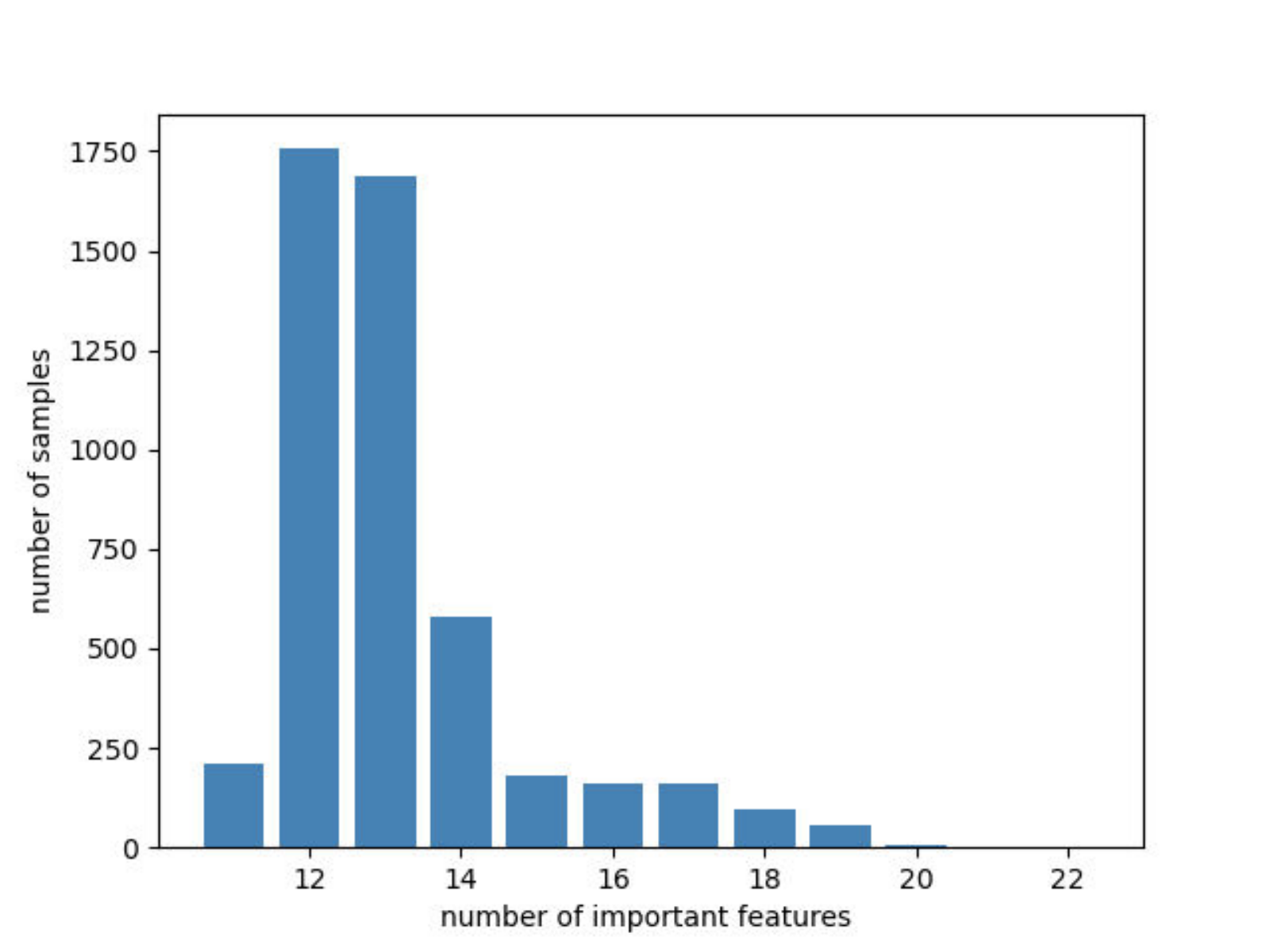}
		}
		\subfigure[\emph{breast}]{
			\includegraphics[width=3.6cm]{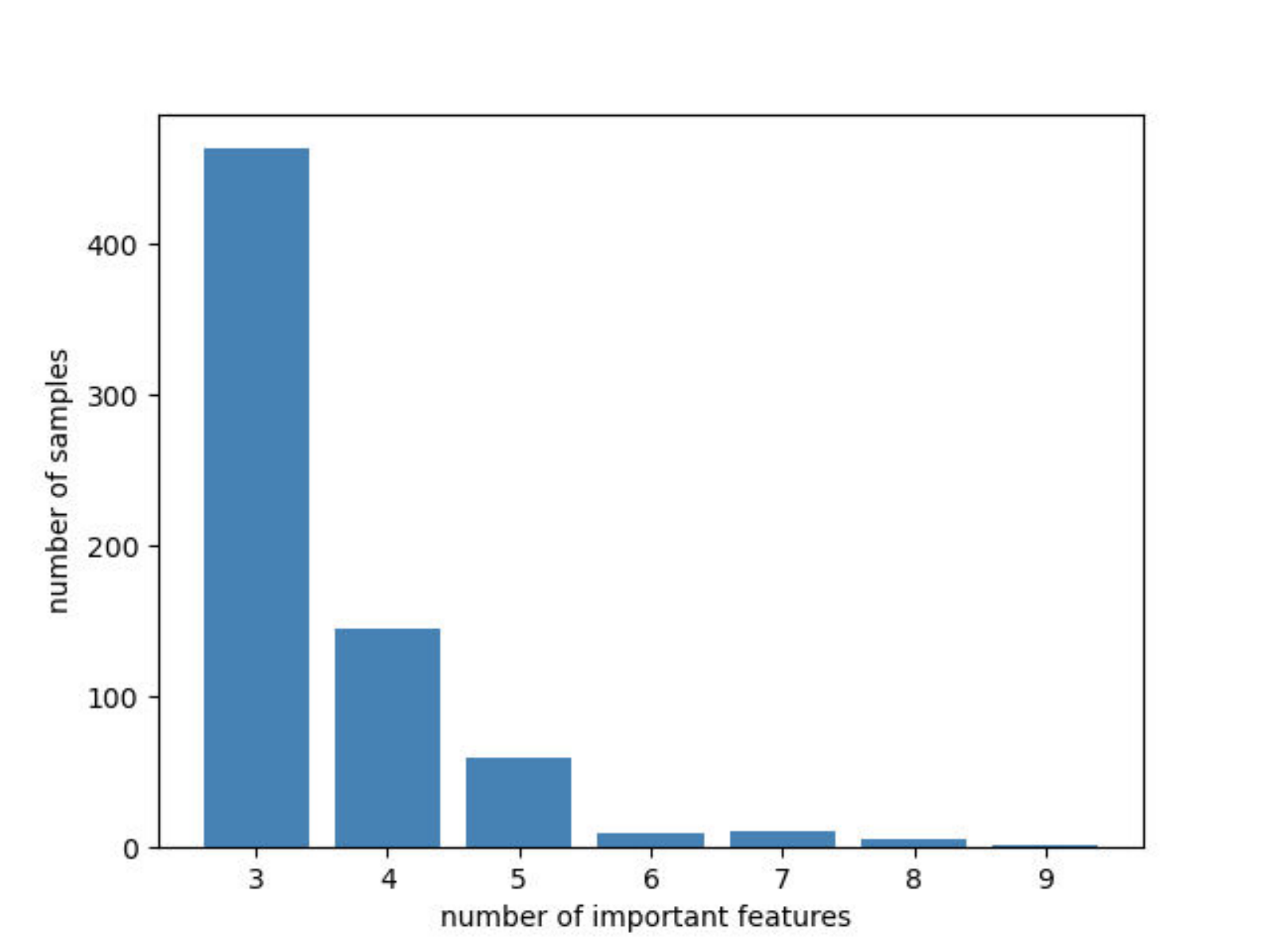}
		}
		\subfigure[\emph{heart}]{
			\includegraphics[width=3.6cm]{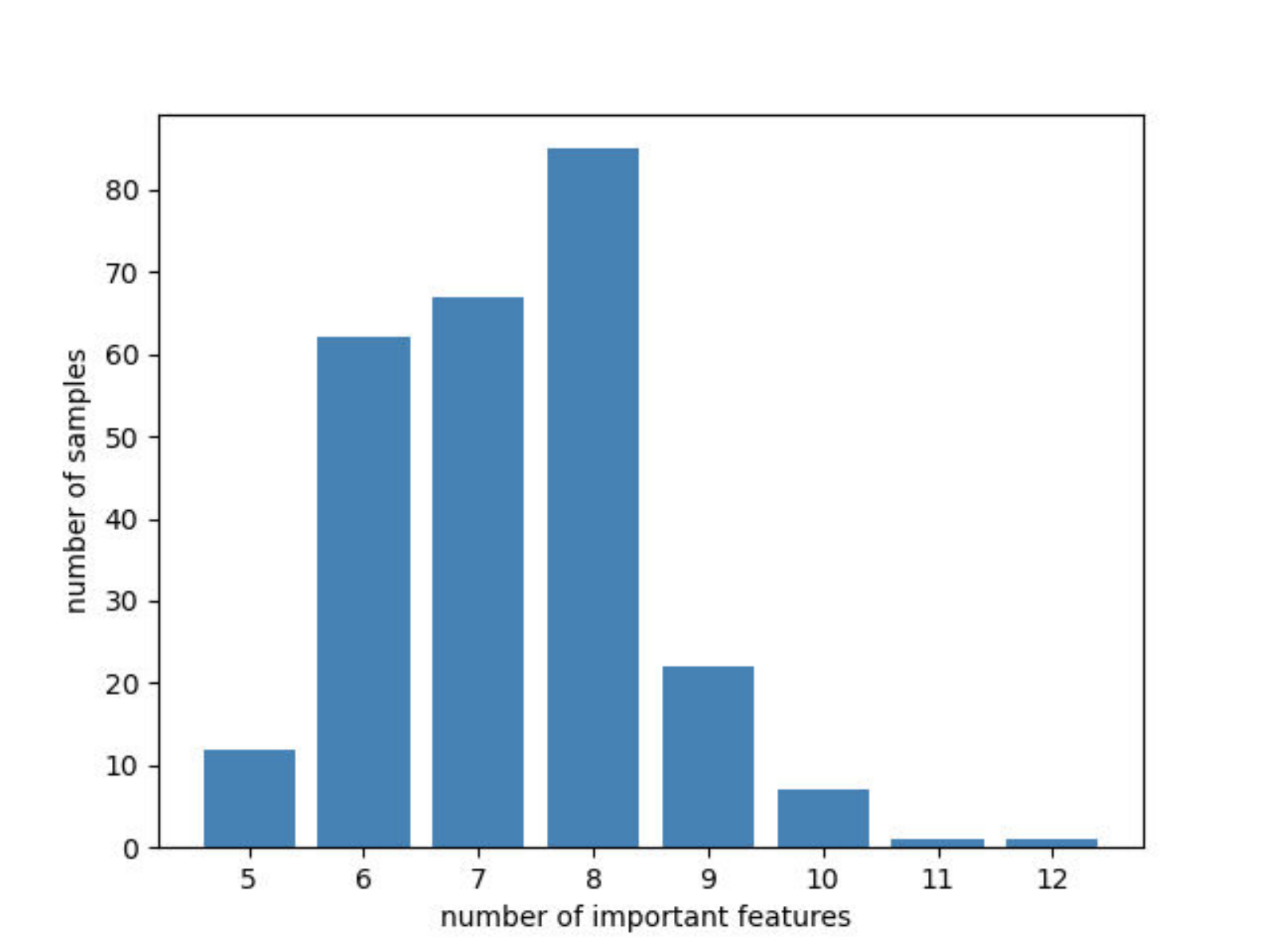}
		}
		\subfigure[\emph{MNIST}]{
			\includegraphics[width=3.6cm]{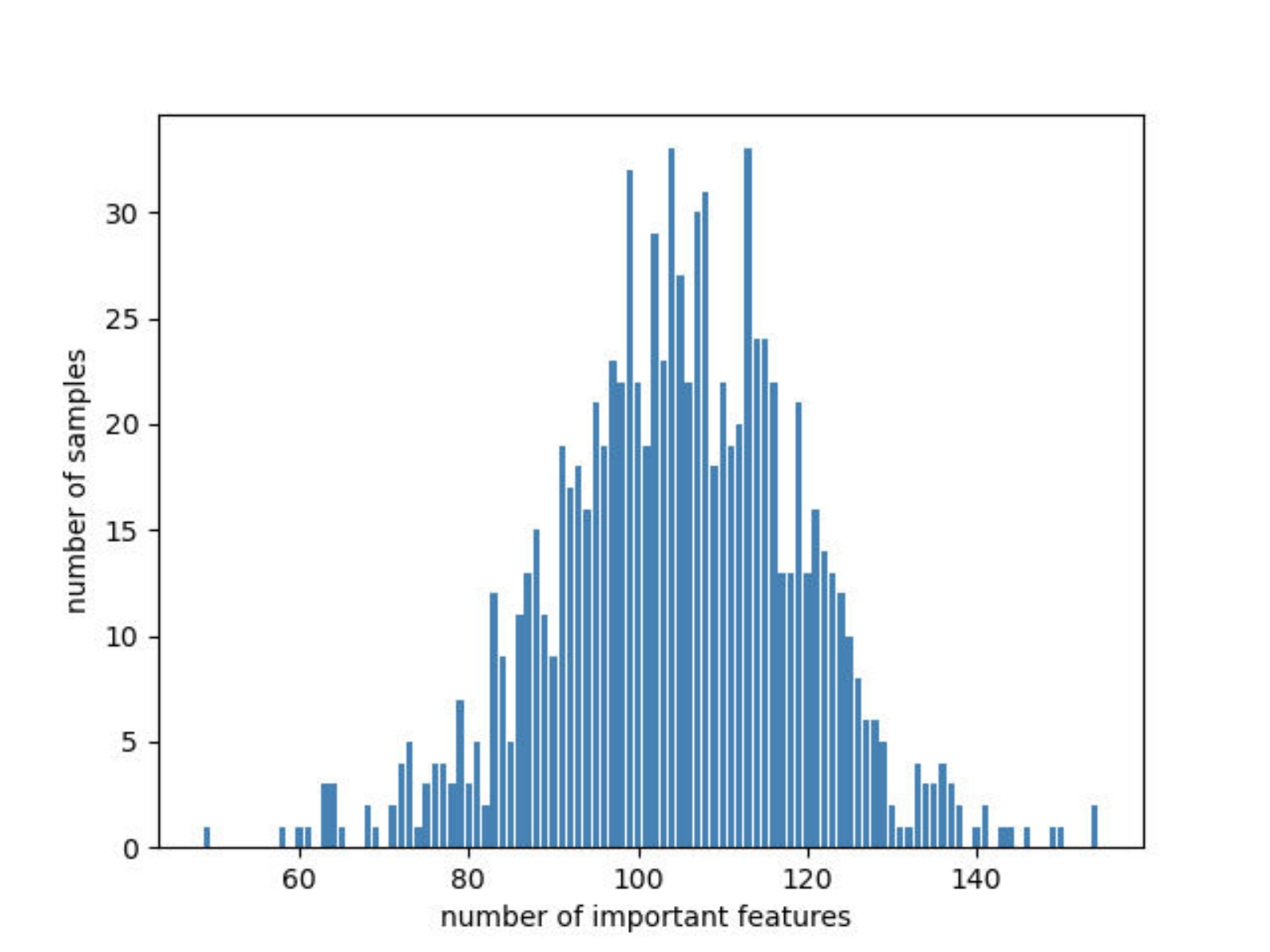}
		}
		\caption{The distribution of the number of important features over testing samples. }
		\label{fig:distribution}
	\end{figure}

\begin{figure}[h]
		\centering
		\subfigure{\includegraphics[width=2.4cm]{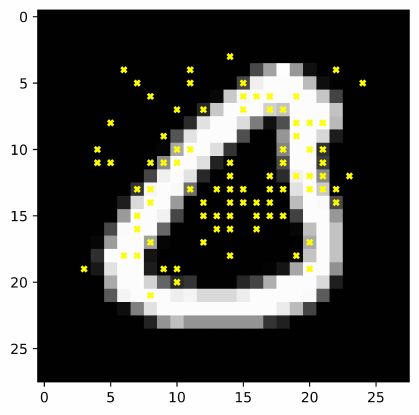}} \hspace{8mm}
		\subfigure{\includegraphics[width=2.4cm]{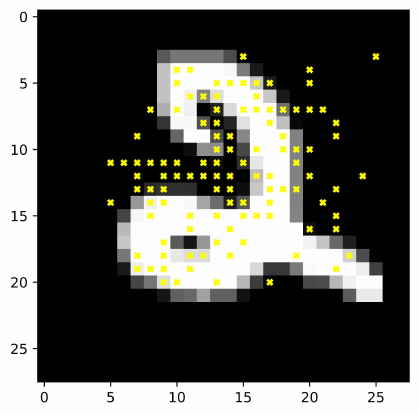}} \hspace{8mm}
		\subfigure{\includegraphics[width=2.4cm]{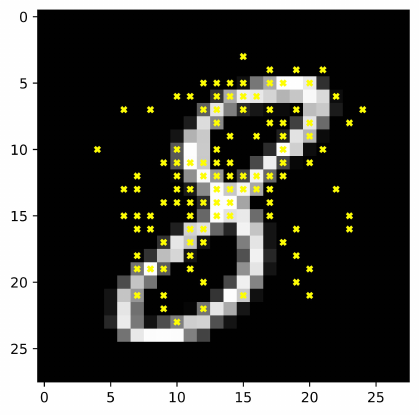}} \\
		\subfigure{\includegraphics[width=2.4cm]{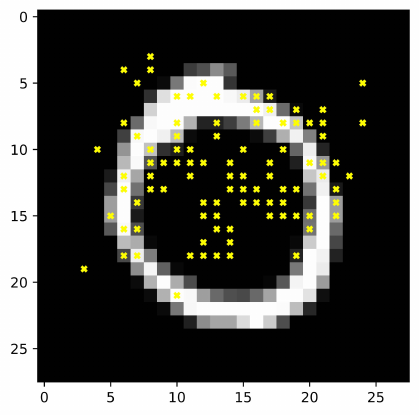}} \hspace{8mm}
		\subfigure{\includegraphics[width=2.4cm]{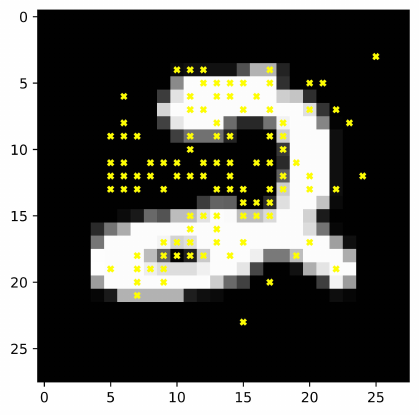}} \hspace{8mm}
		\subfigure{\includegraphics[width=2.4cm]{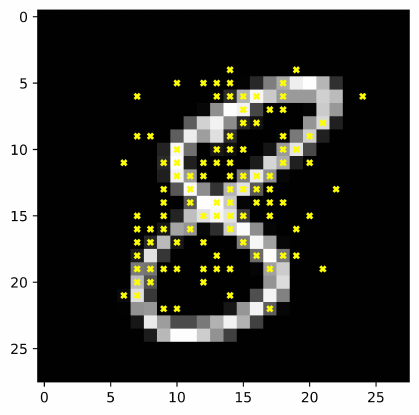}}
		\caption{A visualization of local feature importance on MNIST samples. }
		\label{fig:mnist_explanation}
	\end{figure}

	We evaluate the quality of MUC-driven local explanations on several datasets. We count the number of important features for each test sample, and give a summary of experimental results in Table~\ref{table:results}, where 
	there are the average time of finding importance features (Avg. Time), the number of important features that occur most frequently (Mode), the average number of important features (Avg. Num), the total number of features in the dataset (Total), and the ratio of average / total as feature utilization (Feature Util). Figure \ref{fig:distribution} presents the distribution of the number of important features across all testing samples. Figure \ref{fig:mnist_explanation} visualizes the quality of the explanations on MNIST datasets. The yellow dots locate the important features. 
	
	\noindent\textbf{Discussion.} Table~\ref{table:results} shows that `Mode' and `Avg. Num' numbers are almost the same. Also, the more features there are in the dataset, the longer the time needed for computation. Besides, the maximum feature utilization is 57\%, and the minimum is 13\%, which means that only a small number of features are important in those datasets generally. Figure~\ref{fig:distribution} confirms that the mode is often near the average and that feature utilization is often less than 50\%. From Figure \ref{fig:mnist_explanation} we note that the important features of different images vary, implying the flexibility of our explanation. We also observe that the most influential features spread around the shape of the corresponding digit. Also, the pixels at the corners are unimportant, as expected. These results make sense and align well with our intuitions. In summary, our MUC-driven local explanations are of high quality.
	
	\vspace{-1em}
	
	\begin{table}[h!]
		\caption {Experimental results on our MUC-based method for local feature importance.}
		\centering
		\begin{tabular}{|c|c|c|c|c|c|c|}
			\hline
			Dataset &  lending & bank loan & credit & heart & breast & MNIST\\
			\hline
			Avg. Time &16.23 s  &1.98 s & 14.37 s& 0.68 s& 0.53 s& 70.41 s \\
			\hline
			Mode &12  &3  & 12 & 8 & 3& 104\\
			\hline
			Avg. Num &12 & 3& 13 & 7 & 4& 105\\
			\hline
			Total & 30 & 11 &23 & 13 &9& 784\\
			\hline
			Feature Util & 40\% & 28\%& 57\%& 54\%& 44\%&13\%\\
			\hline
		\end{tabular}
		\label{table:results}
	\end{table}
	
	\vspace{-3em}
	\subsubsection{Global feature importance.}
\begin{figure}[htbp]
		\centering
		\subfigure[\emph{heart, $M = 2 ^ {12}$}]{
			{\includegraphics[width=5.3cm]{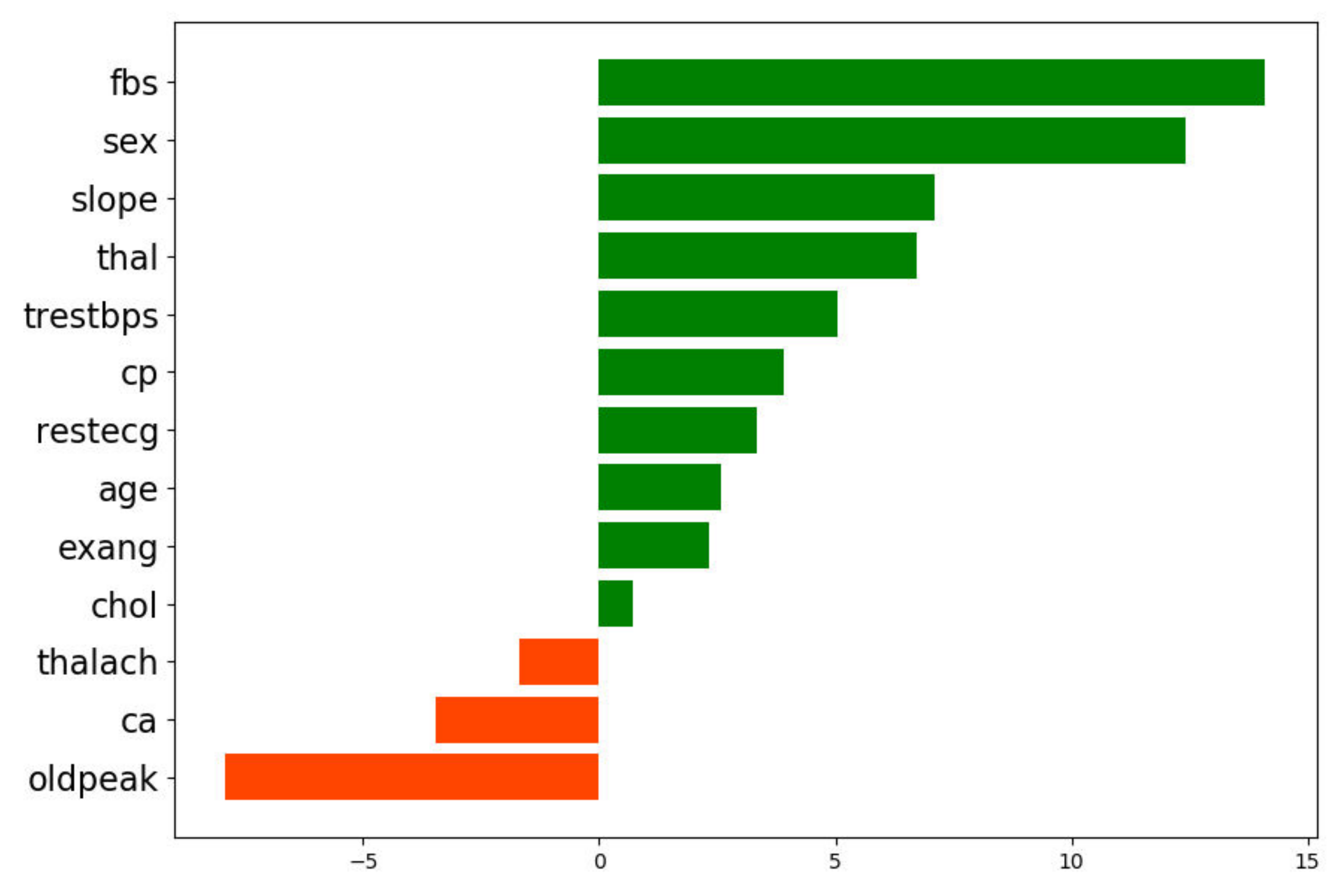}}
		}
		\subfigure[\emph{breast, $M = 2 ^ {8}$}]{
			{\includegraphics[width=5.3cm]{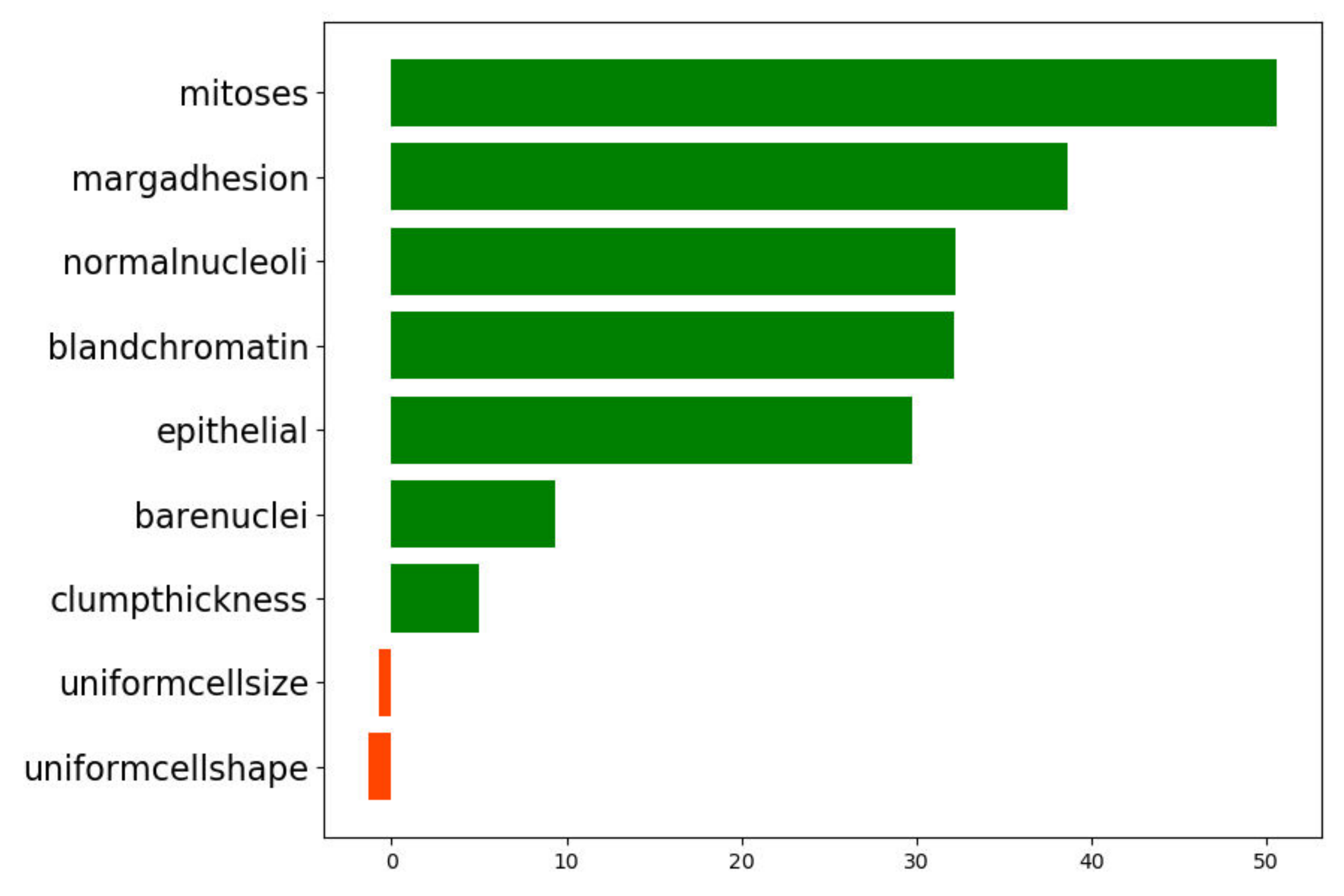}}
		}
		\caption{A visualization of MUC-driven Shapley (M-Shapley) values on models trained with the heart and breast dataset.}
		\label{fig:shap}
	\end{figure}

	Figure~\ref{fig:shap} demonstrates global explanations based on MUC-driven Shapley values (M-Shapley) for the negative class of the heart and breast dataset. The features’ names are listed on the left, and their contributions are sorted and plotted as horizontal bars. Green bars denote the positive impact on the model's prediction, and red ones denote the negative impact. When calculating the importance, all subsets of feature set are visited. Such plots can be used in applications such as medical diagnosis.
	\vspace{-0.2em}
	
	\begin{figure}
		\centering
		\includegraphics[width=10cm]{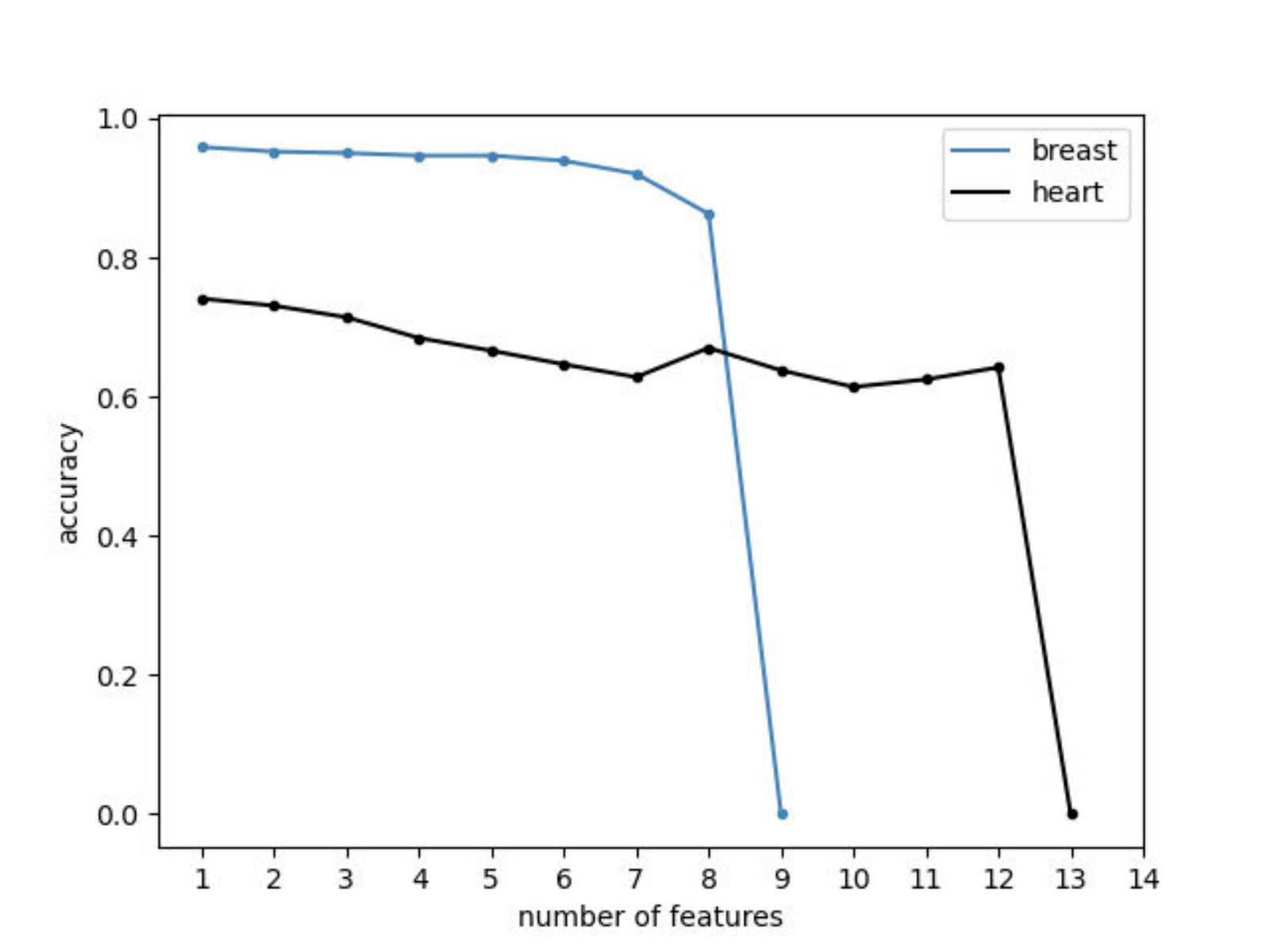}
		\caption{Accuracy of retrained random forest with top N feature values set as mean.}
		\label{fig:accuracy}
	\end{figure}
	
	\noindent\textbf{Effectiveness of M-Shapley.} To evaluate the effectiveness of M-Shapley, we retrain the Random Forest with features selected according to the M-Shapley feature importance ranking and observe the change of the accuracy. Instead of removing features, we retain the model by replacing corresponding values with per feature mean. This evaluation method maintains the consistency of distribution in the training data and the retraining data \cite{hooker2018benchmark}. To eliminate the factor of chance, we run the test on randomly shuffled datasets 30 times. Therefore, the shown accuracy is the average over all results. Figure \ref{fig:accuracy} shows the accuracy vs. top N features observed. As shown, the accuracy keeps decreasing with more informative feature values set as mean, which demonstrates that MUC-driven Shapley values are an effective feature importance measurement.
	
	\subsubsection{Adversarial analysis.}
	We record and compare the total time for generating optimized adversarial samples by the Opt-attack and our method that combines Opt-attack with MUC (MUC-attack). The total time includes the time for determining the initial \(\boldsymbol{\theta}_0\) using fine-grained and binary search algorithm (Search Time) and the time for optimizing (Opt Time). It is worth noting that the search time of the MUC-attack is decomposed into the time to define the adversarial region and that to determine \(\boldsymbol{\theta}_0\). Meanwhile, `Distance' defined by $\frac{1}{m}\sum_{j=1}^{m}\left \| \boldsymbol{x}_j^{org} - \boldsymbol{x}^{*}_j\right \|$ is also taken into account. It represents the average of all distances from original samples $\boldsymbol{x}_j^{org}$ to their corresponding nearest adversarial samples $\boldsymbol{x}^{*}_j$. The parameters in the optimization part of the two methods are set to be the same. Also, for balancing the search time and the search accuracy of Opt-attack, we set the number of candidate adversarial samples to be the minimum of 1000 (suggested in Opt-attack) and the number of samples of other classes in the dataset.
	\begin{table}
		\caption {Experimental results on our adversarial analysis method MUC-attack.}
		\centering
		\begin{tabular}{|c|c|c|c|c|c|}
			\hline
			Dateset & Attack & Search Time & Opt Time & Total Time & Distance \\
			\hline
			\multirow{2}{*}{\emph{lending}}
			& Opt-attack &36.77 s &9.92 s &46.69 s &14943.58 \\
			& MUC-attack &39.97 s &7.95 s &47.92 s & $\boldsymbol{8312.92}$ \\
			\hline
			\multirow{2}{*}{\emph{bank loan}}
			& Opt-attack &25.15 s &12.46 s &37.61 s &25.46 \\
			& MUC-attack &27.92 s &10.55 s &38.48 s &$\boldsymbol{7.75}$ \\
			\hline
			\multirow{2}{*}{\emph{credit}}
			& Opt-attack &101.122 s &59.09 s &160.22 s &23522.53 \\
			& MUC-attack &74.77 s &11.89 s &86.66 s &$\boldsymbol{628.82}$ \\
			\hline
			\multirow{2}{*}{\emph{heart}}
			& Opt-attack &3.69 s &4.22 s &7.92 s &8.38 \\
			& MUC-attack &5.98 s &3.87 s &9.86 s &$\boldsymbol{7.55}$ \\
			\hline
			\multirow{2}{*}{\emph{breast}}
			& Opt-attack &31.32 s &23.64 s &54.97 s &5.66 \\
			& MUC-attack &11.83 s &18.44 s &30.28 s &$\boldsymbol{5.14}$ \\
			\hline
			\multirow{2}{*}{\emph{MNIST}}
			& Opt-attack &4.33 s &65.45 s &69.78 s &387.35 \\
			& MUC-attack &93.29 s &4.91 s &98.2 s &$\boldsymbol{36.53}$ \\
			\hline
		\end{tabular}
		\label{table:adversarial}
	\end{table}
	
	\begin{figure}[ht]
		\centering
		\subfigure[\emph{Raw image}]{
			{\includegraphics[width=2.7cm]{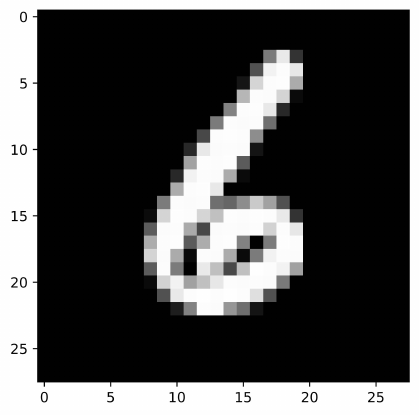}}} \hspace{5mm}
		\subfigure[\emph{MUC-attack}]{
			{\includegraphics[width=2.7cm]{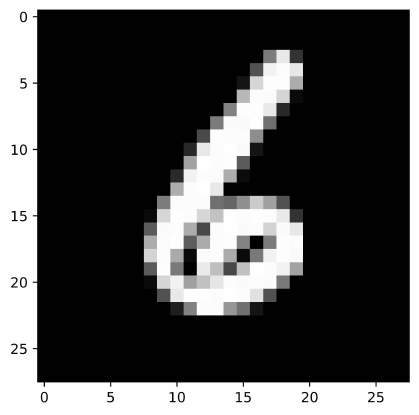}}}
		\hspace{5mm}
		\subfigure[\emph{Opt-attack}]{
			{\includegraphics[width=2.7cm]{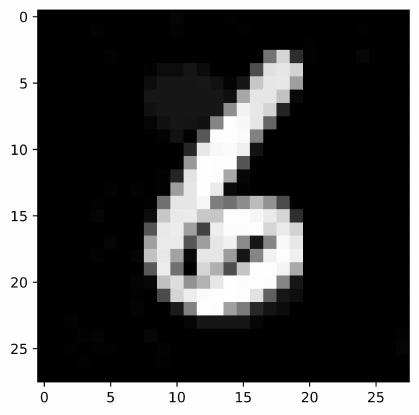}}}
		\caption{Visualization of attacking. The Opt-attack method generates an adversarial sample with visible shades while our method generates a sample that is almost the same as the original.}
		\label{fig:attack}
	\end{figure}
	\noindent\textbf{Discussion.} According to Table \ref{table:adversarial}, the execution time of these two algorithms is roughly the same, while the adversarial samples generated with the guidance of MUC are much closer to the original samples than those without MUC's help because Opt-attack finds fault on every feature. The MUC-attack sacrifices part of the time to identify the adversarial region for better optimization, unlike Opt-attack's unguided search. Figure \ref{fig:attack} shows the raw image and two adversarial images. The subtle perturbations are hardly perceptible to the human eye, while some relatively noticeable differences in light and shade in the Opt-attacked image still exist. Besides, as the initial adversarial sample is selected from a subset of adversarial ones from the original dataset in Opt-attack, the limited sample size gives Opt-attack fewer choices to search closer adversarial samples. In this respect, the MUC-attack performs better. It searches in a continuous space that contains a considerable number of candidate initial adversarial samples. It provides conditions for finding better adversarial samples. Overall, our approach finds closer adversarial samples using a similar time.
	
	\noindent\textbf{Case study.}
	Finally, we will show how the adversarial analysis helps the clients that suffer from rejected loans. We carefully design a user-centered report for them. According to the generated nearest adversarial samples based on Opt-attack and MUC-attack, we list the terms needed to be modified with the extent of the altered values, as shown in table~\ref{table:guidance1} and table~\ref{table:guidance2}. The suggestions in the report provided by MUC-attack do not involve drastic changes to the application and are easier to implement than those by Opt-attack. Thus, the user-centered report based on MUC-attack is more helpful for these clients to improve their applications and eventually obtain approved loans in the future. 
	
	\begin{table}
		\caption {Bank loan application improvements based on MUC-attack.}
		\begin{tabular}{|l|}
			\hline
			Dear No.405415, we provide you the following advice on your application\\
			to help you succeed in loan: \\
			\hline
			Reducing the amount of loan by about 0.7\%; \\
			Increasing the interest rate by 1.6 \%; \\
			Reducing monthly installment by about 0.1\%; \\
			Increasing your income by about 0.18\% \\
			\hline
		\end{tabular}
		\label{table:guidance1}
	\end{table}
	
	\begin{table}
		\caption {Bank loan application improvements based on Opt-attack.}
		\begin{tabular}{|l|}
			\hline
			Dear No.405415, we provide you the following advice on your application\\
			to help you succeed in loan: \\
			\hline
			Reducing the amount of loan by about 5.5\%; \\
			Increasing the interest rate by 21.6 \%; \\
			Reducing monthly installment by about 7.23\%; \\
			Increasing your income by about 4.4\%; \\
			Shorten your loan description by about 8\% \\
			\hline
		\end{tabular}
		\label{table:guidance2}
	\end{table}
	
	\section{Related Work} 
	
	\vspace{-0.8em}
	Related work can be divided into three categories.
	
	First, we set our eyes on explainable ML based on important features. Ribeiro et al. present LIME \cite {ribeiro2016should} to build a local linear model near the instance of interest, and the weights reflect the feature importance. They also present the rule-based Anchor \cite{ribeiro2018anchors}. It extracts IF-THEN rules to explain the predictions. The discretization of numeric features in these two methods causes original information loss. Monotone Influence Measures (MIM) \cite{Sliwinski_Strobel_Zick_2019} use influence functions to express the effect of the feature's value on the likelihood of assigning the label. The influence rules are quite complex and make this method less intuitive. SHAP \cite{lundberg2017unified} produces the important scores on features in game theory. It is for the whole model, not specific to one classification.
	
	Then we review works related to adversarial samples, 
	the application of which on explainable ML is budding in recent years. Pignatiev et al. \cite{ignatiev2019relating} introduce the duality relation between explanations and adversarial samples theoretically. Xu et al. \cite{xu2018structured} regard generated adversarial perturbations as the clear correlations between original and target images, and the adversarial saliency map better interprets the perturbing mechanisms. 
	Most adversarial theories are tailored for neural networks. By contrast, Tolomei et al. \cite{tolomei2017interpretable} propose adversarial analysis specific for Random Forest, but it focuses on the global explanation only.
	
	Another line of research concerns applying formal methods to ML. Ehlers \cite{ehlers2017formal} employs an SMT solver to verify the linear approximation of feed-forward neural networks. Tran et al. \cite{tran2019star} use star-based reachability algorithms to verify the safety and robustness property of Deep Neural Networks. Zhang et al. \cite{songfu2021cav} propose a more scalable tool based on Binary Decision Diagrams to verify the Binarized Neural Networks.  
	Nie et al. \cite{nie2020varf} and Chen el al. \cite{chen2019robustness} verify the robustness of tree ensembles. Moreover, Slias \cite{bride2021silas} analyses the general decision-making through extracting maximum satisfiable core (MSC) of Random Forest in the form of a logical formula. 
	Different from the above, the novelty of our work lies in explaining Random Forest's prediction with logical reasoning.
	
	\section {Conclusion and Future Work} 
	   This paper proposes a comprehensive approach to interpreting Random Forest(RF) through MUC-based logical analysis. We use the minimal unsatisfiable core (MUC) produced by a SMT solver from the encoded decision process of RF to guide the extraction of important features in individual prediction. Based on the logical information uncovered by MUC, we construct MUC-driven Shapley values to measure  feature importance when the model predicts a certain class. Moreover, by leveraging the adversarial region divided by MUC, we propose a novel adversarial analysis that can generate more desirable adversarial samples. Experimental results show that our method provides high-quality and effective feature-relevant explanations. In a comparable time, our proposed adversarial analysis is able to generate closer adversarial samples compared with a state-of-the-art method. In a case study, we show the applicability of our adversarial analysis in providing easy-to-implement suggestions.
	
	
		

	
	\bibliographystyle{splncs04}
	\bibliography{ref}
\end{document}